\pdfoutput=1

\documentclass[12pt]{article}
\usepackage[utf8]{inputenc}
\usepackage[margin=1in]{geometry}
\usepackage{tabularx} 
\usepackage{booktabs}
\usepackage{longtable}
\usepackage{multirow}
\usepackage{float}
\usepackage{siunitx}
\usepackage{hyperref}
\usepackage{url}
\usepackage{amsmath, amssymb}
\usepackage[numbers]{natbib}
\usepackage{ragged2e}
\usepackage[affil-it]{authblk}
\usepackage{graphicx}
\usepackage[T1]{fontenc}   
\usepackage[utf8]{inputenc}
\usepackage{lmodern}       
\usepackage{newunicodechar}
\usepackage{caption}          

\DeclareUnicodeCharacter{2013}{--}   
\DeclareUnicodeCharacter{2014}{---}  
\DeclareUnicodeCharacter{2011}{\mbox{-}} 
\DeclareUnicodeCharacter{00A0}{~}    

\DeclareUnicodeCharacter{2018}{`}    
\DeclareUnicodeCharacter{2019}{'}    
\DeclareUnicodeCharacter{201C}{``}   
\DeclareUnicodeCharacter{201D}{''}   

\DeclareUnicodeCharacter{03BC}{\ensuremath{\mu}}

\title{PARROT: An Open Multilingual Radiology Reports Dataset}
\author[1,2]{Bastien Le Guellec\thanks{Corresponding author: \texttt{bastien.leguellec@chu-lille.fr}}}
\author[3]{Kokou Adambounou}
\author[4]{Lisa C.~Adams}
\author[5]{Thibault Agripnidis}
\author[6]{Sung Soo Ahn}
\author[7]{Radhia Ait Chalal}
\author[8,9]{Tugba Akinci D'Antonoli}
\author[10,11]{Philippe Amouyel}
\author[12]{Henrik Andersson}
\author[10,11]{Raphaël Bentegeac}
\author[13]{Claudio Benzoni}
\author[14]{Antonino Andrea Blandino}
\author[4]{Felix Busch}
\author[15]{Elif Can}
\author[16]{Riccardo Cau}
\author[17]{Armando Ugo Cavallo}
\author[18]{Christelle Chavihot}
\author[19]{Erwin Chiquete}
\author[20]{Renato Cuocolo}
\author[21]{Eugen Divjak}
\author[21]{Gordana Ivanac}
\author[22]{Barbara Dziadkowiec-Macek}
\author[23]{Armel Elogne}
\author[24]{Salvatore Claudio Fanni}
\author[25]{Carlos Ferrarotti}
\author[26]{Claudia Fossataro}
\author[27]{Federica Fossataro}
\author[28]{Katarzyna Fułek}
\author[29]{Michał Fułek}
\author[30,31]{Paweł Gać}
\author[32]{Martyna Gachowska}
\author[33]{Ignacio García-Juárez}
\author[34]{Marco Gatti}
\author[35]{Natalia Gorelik}
\author[36]{Alexia Maria Goulianou}
\author[10,11]{Aghiles Hamroun}
\author[37]{Nicolas Herinirina}
\author[38]{Quentin Holay}
\author[39,40]{Felipe Kitamura}
\author[41]{Michail E.~Klontzas}
\author[42]{Anna Kompanowska}
\author[43]{Rafał Kompanowski}
\author[32]{Krzysztof Kraik}
\author[32]{Dominik Krupka}
\author[1]{Alexandre Lefèvre}
\author[4]{Tristan Lemke}
\author[44]{Maximilian Lindholz}
\author[45]{Lukas Müller}
\author[29,46]{Piotr Macek}
\author[4]{Marcus Makowski}
\author[47]{Luigi Mannacio}
\author[48]{Aymen Meddeb}
\author[47]{Antonio Natale}
\author[18]{Béatrice Nguema Edzang}
\author[49]{Adriana Ojeda}
\author[6]{Yae Won Park}
\author[34]{Federica Piccione}
\author[47]{Andrea Ponsiglione}
\author[50]{Małgorzata Poręba}
\author[30,50]{Rafał Poręba}
\author[13]{Philipp Prucker}
\author[1,2]{Jean-Pierre Pruvo}
\author[14,51]{Rosa Alba Pugliesi}
\author[52]{Feno Hasina Rabemanorintsoa}
\author[53]{Vasileios Rafailidis}
\author[28]{Katarzyna Resler}
\author[32]{Jan Rotkegel}
\author[16]{Luca Saba}
\author[54]{Ezann Siebert}
\author[47]{Arnaldo Stanzione}
\author[55]{Ali Fuat Tekin}
\author[34]{Liz Toapanta-Yanchapaxi}
\author[36]{Matthaios Triantafyllou}
\author[53]{Ekaterini Tsaoulia}
\author[36]{Evangelia Vassalou}
\author[14]{Federica Vernuccio}
\author[12]{Johan Wassélius}
\author[56]{Weilang Wang}
\author[57]{Szymon Urban}
\author[57,58]{Adrian Włodarczak}
\author[57,58]{Szymon Włodarczak}
\author[30]{Andrzej Wysocki}
\author[44]{Lina Xu}
\author[28]{Tomasz Zatoński}
\author[56]{Shuhang Zhang}
\author[4]{Sebastian Ziegelmayer}
\author[1,2]{Grégory Kuchcinski}
\author[4]{Keno K.~Bressem}

\affil[1]{Department of Neuroradiology, Lille University Hospital, Salengro Hospital, Lille, France}
\affil[2]{U1172 LilNCog, Lille Neuroscience \& Cognition, Université Lille, Lille, France}
\affil[3]{Campus University Hospital Centre, Department of Radiology \& Medical Imaging, Lomé, Togo}
\affil[4]{Department of Diagnostic \& Interventional Radiology, Klinikum rechts der Isar, TUM University Hospital, Technical University of Munich, Munich, Germany}
\affil[5]{Interventional Radiology, University Hospital Timone (AP-HM), Marseille, France}
\affil[6]{Department of Radiology \& Research Institute of Radiological Science \& Center for Clinical Imaging Data Science, Yonsei University College of Medicine, Seoul, South Korea}
\affil[7]{Department of Radiology, Bab El-Oued University Hospital, Algiers, Algeria}
\affil[8]{Diagnostic \& Interventional Neuroradiology, University Hospital Basel, Basel, Switzerland}
\affil[9]{Pediatric Radiology, University Children's Hospital Basel, Basel, Switzerland}
\affil[10]{U1167 RID-AGE, Pasteur Institute of Lille, Inserm, Lille University, Lille, France}
\affil[11]{Public Health – Epidemiology, Lille University Hospital Center, Lille, France}
\affil[12]{Medical Imaging \& Physiology, Skåne University Hospital, Lund, Sweden}
\affil[13]{Institute of AI \& Informatics in Medicine (AIIM), TUM University Hospital, Technical University of Munich, Munich, Germany}
\affil[14]{Radiology, Department of Biomedicine, Neuroscience \& Advanced Diagnostics (BiND), University of Palermo, Palermo, Italy}
\affil[15]{Diagnostic \& Interventional Radiology, Medical Center – University of Freiburg, Faculty of Medicine, Freiburg, Germany}
\affil[16]{Radiology, Azienda Ospedaliero-Universitaria (A.O.U.) di Cagliari, Monserrato, Cagliari, Italy}
\affil[17]{Division of Radiology, Istituto Dermopatico dell'Immacolata (IDI) IRCCS, Rome, Italy}
\affil[18]{Department of Radiology, Hôpital Instruction des Armées, Libreville, Gabon}
\affil[19]{Department of Neurology, Instituto Nacional de Ciencias Médicas y Nutrición Salvador Zubirán, Mexico City, Mexico}
\affil[20]{Department of Medicine, Surgery \& Dentistry, University of Salerno, Baronissi, Italy}
\affil[21]{Department of Diagnostic \& Interventional Radiology, University Hospital Dubrava, Zagreb, Croatia}
\affil[22]{Department of Physiology and Pathophysiology, Wroclaw Medical University, Wroclaw, Poland}
\affil[23]{Department of Radiology, Hôpital Militaire d'Abidjan, Abidjan, Côte d'Ivoire}
\affil[24]{Department of Translational Research, Academic Radiology, University of Pisa, Pisa, Italy}
\affil[25]{Department of Diagnostic Imaging, CEMIC “Norberto Quirno”, Buenos Aires, Argentina}
\affil[26]{Department of Ophthalmology, Catholic University “Sacro Cuore”, Rome, Italy}
\affil[27]{Department of Ophthalmology, ASST Fatebenefratelli Sacco, Milan, Italy}
\affil[28]{Department and Clinic of Otolaryngology, Head and Neck Surgery, Wroclaw Medical University, Wroclaw, Poland}
\affil[29]{Department and Clinic of Diabetology, Hypertension and Internal Diseases, Institute of Internal Diseases, Wroclaw Medical University, Wroclaw, Poland}
\affil[30]{Department of Radiology and Diagnostic Imaging, 4th Military Hospital, Wroclaw, Poland}
\affil[31]{Department of Environmental Health, Occupational Medicine and Epidemiology, Wroclaw Medical University, Wroclaw, Poland}
\affil[32]{Faculty of Medicine, Wroclaw Medical University, Wroclaw, Poland}
\affil[33]{Department of Gastroenterology, Instituto Nacional de Ciencias Médicas y Nutrición Salvador Zubirán, Mexico City, Mexico}
\affil[34]{Department of Surgical Sciences, Radiology Unit, University of Turin, Turin, Italy}
\affil[35]{Department of Radiology, McGill University Health Center, Montreal, Quebec, Canada}
\affil[36]{Department of Medical Imaging, University Hospital of Heraklion, Heraklion, Greece}
\affil[37]{Department of Radiology, Tanambao University Hospital, Antsiranana, Madagascar}
\affil[38]{Department of Radiology, Sainte-Anne Teaching Military Hospital, Toulon, France}
\affil[39]{Bunkerhill Health, San Francisco, CA, USA}
\affil[40]{Department of Diagnostic Imaging, Universidade Federal de São Paulo (UNIFESP), São Paulo, Brazil}
\affil[41]{AI \& Translational Imaging Lab, Department of Radiology, University of Crete, Heraklion, Greece}
\affil[42]{Department of Pediatrics, Klodzko County Hospital, Klodzko, Poland}
\affil[43]{Orthopedics and Traumatology Department of the Musculoskeletal System, Specialist Medical Centre, Polanica-Zdrój, Poland}
\affil[44]{Department of Radiology, Charité University Hospital Berlin, Berlin, Germany}
\affil[45]{Department of Diagnostic \& Interventional Radiology, University Medical Center – Johannes Gutenberg-University Mainz, Mainz, Germany}
\affil[46]{Department of Cardiology, Marciniak Lower Silesian Specialist Hospital, Wroclaw, Poland}
\affil[47]{Department of Advanced Biomedical Sciences, University of Naples Federico II, Naples, Italy}
\affil[48]{Department of Neuroradiology, Charité University Hospital Berlin, Berlin, Germany}
\affil[49]{Departamento de Neuroradiología, DMO, Rosario, Argentina}
\affil[50]{Department of Biological Principles of Physical Activity, Wroclaw University of Health and Sport Sciences, Wroclaw, Poland}
\affil[51]{Division of Health Care Sciences, Dresden International University, Dresden, Germany}
\affil[52]{Department of Radiology, Morafeno Toamasina University Hospital, Toamasina, Madagascar}
\affil[53]{Department of Radiology, Aristotle University of Thessaloniki, AHEPA University General Hospital, Thessaloniki, Greece}
\affil[54]{Department of Ophthalmology, Sir Charles Gairdner Hospital, Perth, Australia}
\affil[55]{Department of Radiology, Başakşehir Çam \& Sakura City Hospital, Istanbul, Turkey}
\affil[56]{Department of Radiology, Zhongda Hospital, Southeast University, Nanjing, China}
\affil[57]{Department of Cardiology, The Copper Health Center, Lubin, Poland}
\affil[58]{Faculty of Medicine, Wroclaw University of Science and Technology, Wroclaw, Poland}

\begin{document}

\date{}
\maketitle
\newpage

\makeatletter
\renewenvironment{abstract}{%
    \section*{\centering \abstractname} 
    \normalfont\normalsize
    \setlength{\parindent}{0pt}
    \setlength{\leftskip}{0pt}%
    \setlength{\rightskip}{0pt}%
}{\par}
\makeatother

\begin{abstract}

\textbf{Rationale and Objectives:}  
To develop and validate \textbf{PARROT}  (Polyglottal Annotated Radiology Reports for Open Testing), a large, multicentric, open-access dataset of fictional radiology reports spanning multiple languages for testing natural-language-processing applications in radiology.\\[4pt]
\textbf{Materials and Methods:}  
From May to September  2024, radiologists were invited to contribute fictional radiology reports following their standard reporting practices. Contributors provided at least  20 reports with associated metadata including anatomical region, imaging modality, clinical context, and—for non-English reports—English translations. All reports were assigned ICD-10 codes. A human-versus-AI report-differentiation study was conducted with 154 participants (radiologists, healthcare professionals, and non-healthcare professionals) assessing whether reports were human-authored or AI-generated.\\[4pt]
\textbf{Results:}  
The dataset comprises 2\,658 radiology reports from 76 authors across 21 countries and 13 languages. Reports cover multiple imaging modalities (CT  36.1\,\%, MRI  22.8\,\%, radiography  19.0\,\%, ultrasound  16.8\,\%) and anatomical regions, with chest  (19.9\,\%), abdomen  (18.6\,\%), head  (17.3\,\%), and pelvis  (14.1\,\%) being most prevalent. In the differentiation study, participants achieved 53.9\,\% accuracy (95\,\%  CI  50.7,57.1) in distinguishing human from AI-generated reports, with radiologists performing significantly better (56.9\,\%; 95\,\%  CI  53.3,60.6; $p<0.05$) than other groups.\\[4pt]
\textbf{Conclusion:}  
PARROT represents the largest openly available multilingual radiology-report dataset, enabling development and validation of NLP applications across linguistic, geographic, and clinical boundaries without privacy constraints.
\end{abstract}


\section{Introduction}

Radiology reports are the communication bridge between radiologists and clinicians, capturing complex visual findings in actionable text that directly informs patient-care decisions  \cite{hartung2020}. The quality and clarity of these reports significantly impact diagnostic accuracy and treatment outcomes, with communication breakdowns representing a notable source of medical errors  \cite{hawkins2014,nobel2022,schwartz2011}.

Large language models (LLMs) have demonstrated remarkable capabilities in enhancing radiology workflows  \cite{keshavarz2024,gertz2024,leguellec2025}. Recent studies show that these models can automatically structure free-text findings into standardized formats suitable for data sharing  \cite{adams2023,leguellec2024}, simplify complex radiological terminology for non-specialists  \cite{amin2023}, and detect inconsistencies and errors with performance comparable to expert radiologists  \cite{gertz2024}. Such capabilities hold substantial promise for improving communication efficiency and report quality in clinical practice.

However, a considerable barrier to deploying LLMs is their English-language centricity. Widely used datasets such as \textsc{MIMIC-IV} and \textsc{MIMIC-CXR} contain thousands of radiology reports but represent only English-language practice  \cite{johnson2023,johnson2019}. Privacy concerns and stringent data-sharing regulations constrain studies involving non-English radiology reports to proprietary datasets that remain inaccessible to the wider research community  \cite{leguellec2024}. This absence of openly available multilingual datasets undermines scientific reproducibility and comparability across different healthcare systems  \cite{wu2024}.

The consequences of this linguistic bias extend beyond simple language barriers. Radiology reporting practices exhibit substantial diversity worldwide—not just in language but also in reporting styles, preferred formats, terminology, and conventions. These variations are so pronounced that evaluation metrics developed for one region often fail to generalize across reports from different countries  \cite{banerjee2024}. Compounding the issue, even advanced LLMs show performance disparities across languages, with capabilities in low-resource languages lagging behind English  \cite{banerjee2024,alduwais2024}.

Developing truly multilingual resources that reflect authentic native-language radiology practice is therefore essential. Preserving language-specific medical terminology is crucial for clinical accuracy  \cite{campos2017}, while capturing diverse reporting conventions ensures that AI tools align with existing clinical workflows. Furthermore, as AI-generated reports become increasingly prevalent, it is important to know whether human-authored reports can be distinguished from AI text; evaluating models solely on synthetic reports may propagate subtle inaccuracies.

In this context, we introduce \textbf{PARROT}, a collaborative initiative designed to overcome multilingual barriers in radiology-AI research. By bringing together radiologists and AI researchers to curate expert-authored reports from diverse linguistic and clinical environments, PARROT provides an open benchmark for developing inclusive radiology-AI tools that can serve global healthcare needs effectively and equitably.

\section{Methods}

\subsection{Study design and recruitment}
The \textbf{PARROT} initiative ran from May  2024 to September  2024. The project was advertised through social-media platforms, professional radiology societies, and direct networking. To ensure broad global representation—particularly from under-represented regions—we leveraged the COMFORT-AI survey network, which connects with radiologists worldwide  \cite{busch2024}. Because the dataset consists entirely of fictional reports with no patient data, institutional-review-board approval was formally waived. Each contributor attested that all submitted reports were fabricated. The fictional nature of the corpus enables unrestricted sharing, modification, and augmentation under a CC-BY-NC-SA  4.0 licence. Unlike datasets containing real patient data, PARROT can be freely distributed across institutional and national boundaries, shared with cloud-based models, facilitating collaborative research without privacy constraints.

\subsection{Submission requirements}
Contributors had to submit at least 20 fictional reports with no upper limit. Each submission required the complete radiological report following the contributor's typical reporting style for their language and region. Contributors were encouraged to use the format, terminology, and structure they would typically employ in clinical practice to capture authentic stylistic variations. Report metadata included anatomical region, imaging modality (CT, MRI, US, XR, etc.), brief clinical context/indication and report language. For non-English reports, contributors provided an English translation to facilitate cross-linguistic analysis. Contributors were given freedom to choose specialties and clinical scenarios according to their expertise. They were specifically instructed to create plausible but non-specific clinical scenarios, include typical incidental findings appropriate for the demographic, and incorporate normal anatomic variants at realistic frequencies. This approach allowed for capturing authentic reporting styles while ensuring the dataset represents a realistic distribution of normal and pathological findings. 
ICD-10 codes were assigned either by the contributor or by B.\,Le  Guellec with assistance from the \texttt{o3-mini-high} language model (OpenAI). Discrepancies were resolved via dialogue with contributors.

\subsection{Dataset organisation and processing}
The final dataset was organized in a standardized structure to facilitate research use (\texttt{JSONL} file). The entire dataset was version-controlled using Git, with the initial public release tagged as v1.0. The repository includes documentation and usage examples to facilitate adoption. This version control approach enables ongoing collaboration and contribution, allowing the dataset to grow and improve over time while maintaining traceability of changes. Future contributors can submit additional reports following the established protocols, enhancing the linguistic and clinical diversity of the resource. The project is accessible at \url{https://github.com/PARROT-reports/PARROT_v1.0/}.

\subsection{Human,versus,AI report differentiation study}
To validate the authenticity and distinctive characteristics of PARROT reports compared to AI-generated content, we conducted a discrimination study. We generated AI reports using GPT-o1 (OpenAI) and selected a comparable number of PARROT reports for comparison. Participants recruited through social media (LinkedIn) and professional networks (e.g., European Society of Medical Imaging Informatics) were asked to identify whether reports were human-authored or AI-generated. The study was conducted on English, German, Italian, French, Greek and Polish reports via Google Docs, where participants selected their language of practice, reviewed and voted on 10 reports, with 5 being AI-generated and 5 human-authored from PARROT. Reports included MRI, CT, X-ray and echographic examinations from multiple anatomical regions. We specifically targeted a mix of radiologists, computer scientists, and general participants to assess whether professional expertise influenced the ability to distinguish authentic radiological reporting from AI-generated content. 
This evaluation was motivated by concerns that researchers might use language models to generate synthetic radiology reports for testing purposes, potentially introducing subtle but important deviations from authentic clinical documentation. We hypothesized that those reports might be difficult to differentiate from human-generated content, but that domain experts such as radiologists would better recognize these differences than non-experts, highlighting the value of professionally authored content in the PARROT dataset.

\subsection{Statistical analysis}
A descriptive analysis was performed to characterize the dataset. Country of origin and frequency of imaging modalities were summarized, and body regions were categorized to assess the diversity of the contributed scenarios. Preliminary maps of geographic origins were generated to visualize the global participation in PARROT. All statistical analyses and visualizations were carried out using R version 4.3.1. Key analytical steps included summarizing submission counts per country and language, enumerating modality usage across the dataset, grouping body regions under standardized terms, analyzing the distribution of normal versus abnormal findings, and categorizing reports by ICD-10 classification. The analysis focused on demonstrating the breadth and diversity of the dataset across linguistic, geographic, and clinical dimensions to establish its utility as a representative multilingual resource for radiology AI research. Differences in performance for the differentiation study were assessed with a $\chi^{2}$ test, with a significance level of $\alpha = 0.05$.

\section{Results}

\subsection{Report origin and linguistic diversity}

The PARROT dataset comprises \num{2658} fictional radiology reports contributed by radiologists in 21 countries across four continents. Poland provided the largest share (837\,reports, 31.5 \%), followed by Germany (316; 11.9 \%), Italy (285; 10.7 \%), Croatia (200; 7.5 \%), and France (173; 6.5 \%).  Figure~\ref{fig:1} illustrates the geographic distribution, which is dominated by Europe but includes growing participation from Asia, South America, and Africa.  Nations from the Global South—Argentina (71; 2.7 \%), China (100; 3.8 \%), and Mexico (75; 2.8 \%)—together represent roughly 19 \% of the collection.

The PARROT dataset encompasses reports in 13 different languages. The distribution of languages reflects the geographical representation, with Polish, German, Italian, and French being the most prevalent. Notably, French-language reports originated from multiple countries including Algeria, France, Ivory Coast, Madagascar, Togo, and Canada, highlighting regional variations in reporting practices within the same language. Similarly, Spanish-language reports were contributed from both Argentina and Mexico, capturing different regional medical terminologies and reporting conventions.

\subsection{Report content and structure}

\begin{figure}[H]
    \centering
    \includegraphics[width=\textwidth,trim=0cm 5cm 0cm 0cm,clip,page=1]{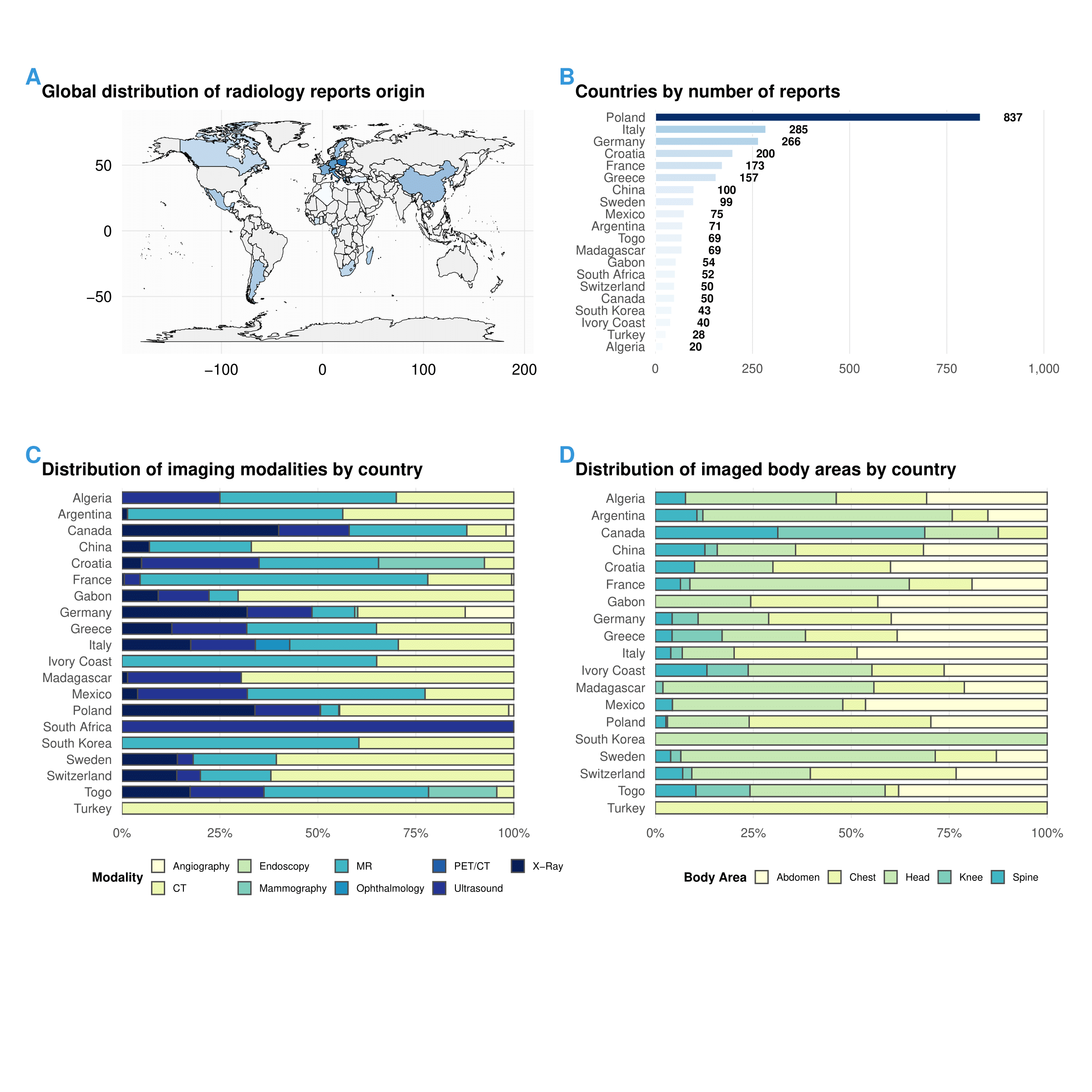} 
    \caption{Geographic distribution and characteristics of the PARROT dataset. A) World map showing the distribution of radiology reports by country of origin, with darker blue indicating higher numbers of reports. B) Bar chart displaying the count of reports by country, arranged in descending order. C) Distribution of imaging modalities by country, showing the percentage of each modality within countries. D) Distribution of five most imaged body areas by country, revealing regional preferences in anatomical focus.}
    \label{fig:1}
\end{figure}

Analysis of report content revealed diverse imaging modalities, with computed tomography (CT) representing \(36.1\% \,(n = 989)\) of the dataset, followed by magnetic resonance imaging (MRI) (\(22.8\%,\, n = 625\)), conventional radiography (XR, \(19.0\%,\, n = 520\)), and ultrasound (US, \(16.8\%,\, n = 461\)) as displayed in Figure~\ref{fig:1}.

The anatomical distribution analysis demonstrated a predominance of reports covering chest (\(n = 677,\, 19.9\%\)), abdomen (\(n = 631,\, 18.6\%\)), head (\(n = 588,\, 17.3\%\)), and pelvis (\(n = 480,\, 14.1\%\)) as detailed in Table~\ref{tab:bodyareas} and visualized in Figure~\ref{fig:2}. Less-commonly represented regions included extremities and specialized areas such as the orbit or pituitary gland. The relationship between modalities and anatomical regions is shown in Supplementary Figure S1.

The median report length varied across languages. The longest reports were in Turkish (median length \(382\) words; interquartile range, IQR: \(154.75\) words), followed by Spanish (\(196\); IQR \(136.5\)) and French (\(172\); IQR \(130\)), with the shortest reports being in Afrikaans with a median word count of \(36.5\) (IQR \(12.75\)). Further details on word count are provided in Table~\ref{tab:wordcount}.

\begin{table}[H]
  \centering
  \caption{Distribution of body areas in PARROT.}
  \label{tab:bodyareas}
  \begin{tabular}{lrr}
    \toprule
    \textbf{Body Area} & \textbf{Count} & \textbf{Percentage (\%)}\\ \midrule
    Chest & 677 & 19.9\\
    Abdomen & 631 & 18.6\\
    Head & 588 & 17.3\\
    Pelvis & 480 & 14.1\\
    Breast & 176 & 5.2\\
    Neck & 138 & 4.1\\
    Lumbar Spine & 102 & 3.0\\
    Spine & 95 & 2.8\\
    Orbit & 79 & 2.3\\
    Cervical & 71 & 2.1\\
    Knee & 60 & 1.8\\
    Lower Extremity & 50 & 1.5\\
    Shoulder & 49 & 1.4\\
    Thoracic Spine & 46 & 1.4\\
    Upper Extremity & 30 & 0.9\\
    Foot & 22 & 0.6\\
    Wrist & 21 & 0.6\\
    Ankle & 20 & 0.6\\
    Face & 15 & 0.4\\
    Pituitary Gland & 12 & 0.4\\
    Elbow & 10 & 0.3\\
    Testis & 8 & 0.2\\
    Whole Body & 8 & 0.2\\
    Ear & 4 & 0.1\\
    Hip & 3 & 0.1\\
    Nose & 2 & 0.1\\ \bottomrule
  \end{tabular}
\end{table}

\begin{table}[H]
  \centering
  \caption{Median word count of reports by language.}
  \label{tab:wordcount}
  \begin{tabular}{lc}
    \toprule
    \textbf{Language} & \textbf{Median (IQR) words}\\ \midrule
    Afrikaans & 36.5 (12.8)\\
    Chinese & 182.5 (137)\\
    Croatian & 67 (41)\\
    Dutch & 39.5 (15.3)\\
    French & 172 (130)\\
    German & 71.5 (79.3)\\
    Greek & 86 (72)\\
    Italian & 94 (87)\\
    Korean & 43 (28)\\
    Polish & 112 (145)\\
    Spanish & 196 (136.5)\\
    Swedish & 73 (73)\\
    Turkish & 328 (154.8)\\ \bottomrule
  \end{tabular}
\end{table}

\subsection{Human,versus,AI report differentiation study}

\begin{figure}[H]
    \centering
    \includegraphics[width=\textwidth,trim=0cm 0cm 0cm 0cm,clip,page=1]{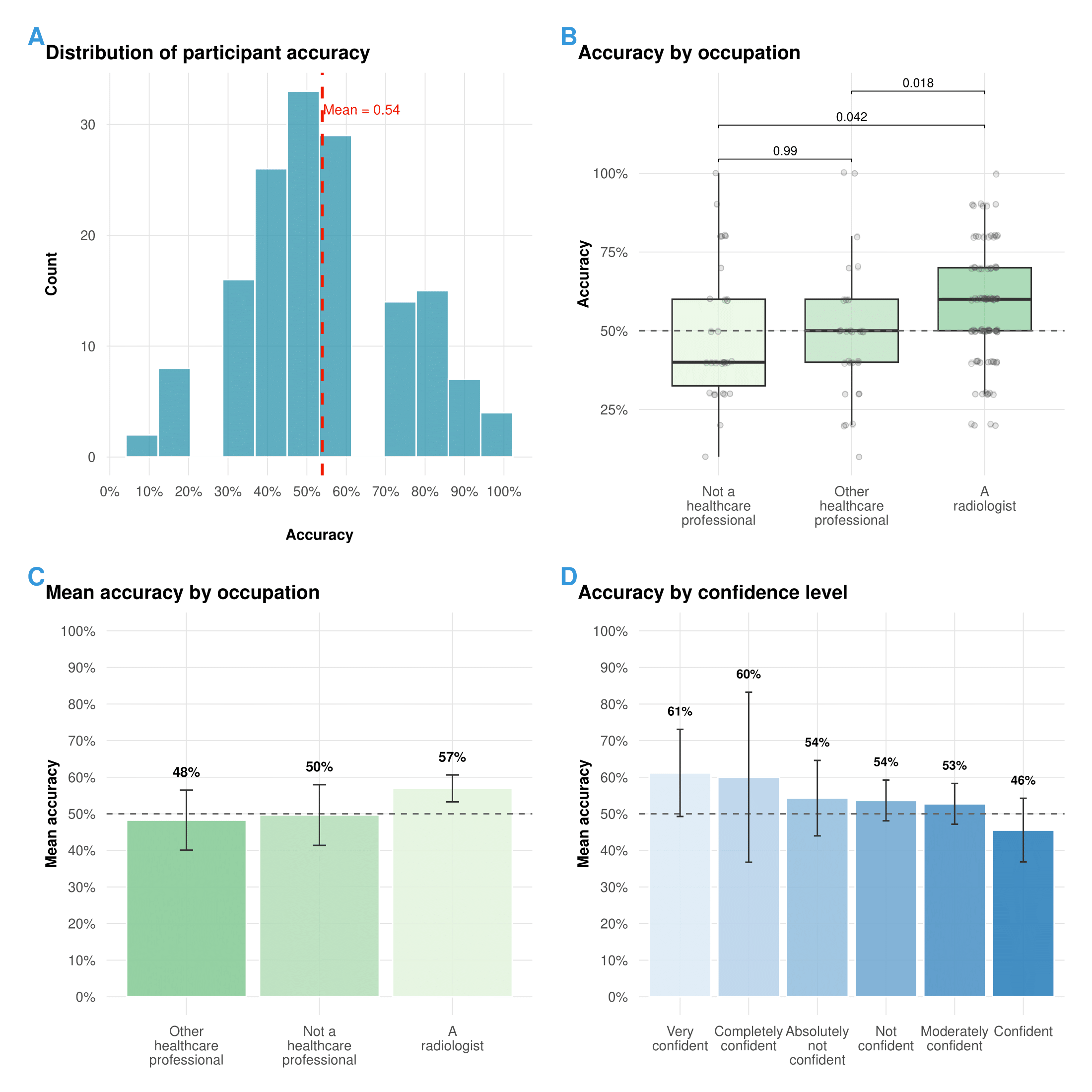} 
    \caption{Results of the human vs. AI report differentiation study A) Distribution of participant accuracy in distinguishing between human-authored and AI-generated radiology reports, with dashed line representing chance level (50\%). B) Box plots showing accuracy by occupation group, with radiologists performing significantly better than other groups. C) Effect sizes from logistic regression model showing factors associated with accuracy in the differentiation task. D) Accuracy by self-reported confidence level, revealing modest correlation between confidence and performance.
}
    \label{fig:2}
\end{figure}

To evaluate the authenticity of the human-authored reports in PARROT compared with AI-generated content, we conducted a differentiation study with \(154\) participants from diverse backgrounds. Overall, participants achieved a mean accuracy of \(53.9\%\) \((95\%~\text{CI}: 50.7\text{--}57.1\%)\) in distinguishing between human-authored and AI-generated reports, indicating performance only slightly above chance level (Figure~\ref{fig:2}).

Analysis by occupation revealed that radiologists performed better (\(56.9\%,\ 95\%~\text{CI}: 53.3\text{--}60.6\%\)) than non-healthcare professionals (\(49.7\%,\ 95\%~\text{CI}: 41.4\text{--}58.0\%\)) and other healthcare professionals (\(48.3\%,\ 95\%~\text{CI}: 40.1\text{--}56.5\%\)), as shown in Figure~\ref{fig:2}. This difference was statistically significant (\(p < 0.05\)), suggesting that domain expertise provides some advantage in discerning report authenticity.

Self-reported confidence showed only a weak association with actual performance. Participants reporting the highest confidence levels (“Very confident’’ and “Completely confident’’) achieved modestly higher accuracy (\(61.2\%\) and \(60.0\%\), respectively) compared with those with lower confidence ratings (Figure~\ref{fig:2}). However, the correlation between confidence and accuracy was relatively weak (\(r = 0.07\)), indicating poor metacognition in this task.

\begin{figure}[H]
    \centering
    \includegraphics[width=\textwidth,trim=0cm 0cm 0cm 0cm,clip,page=1]{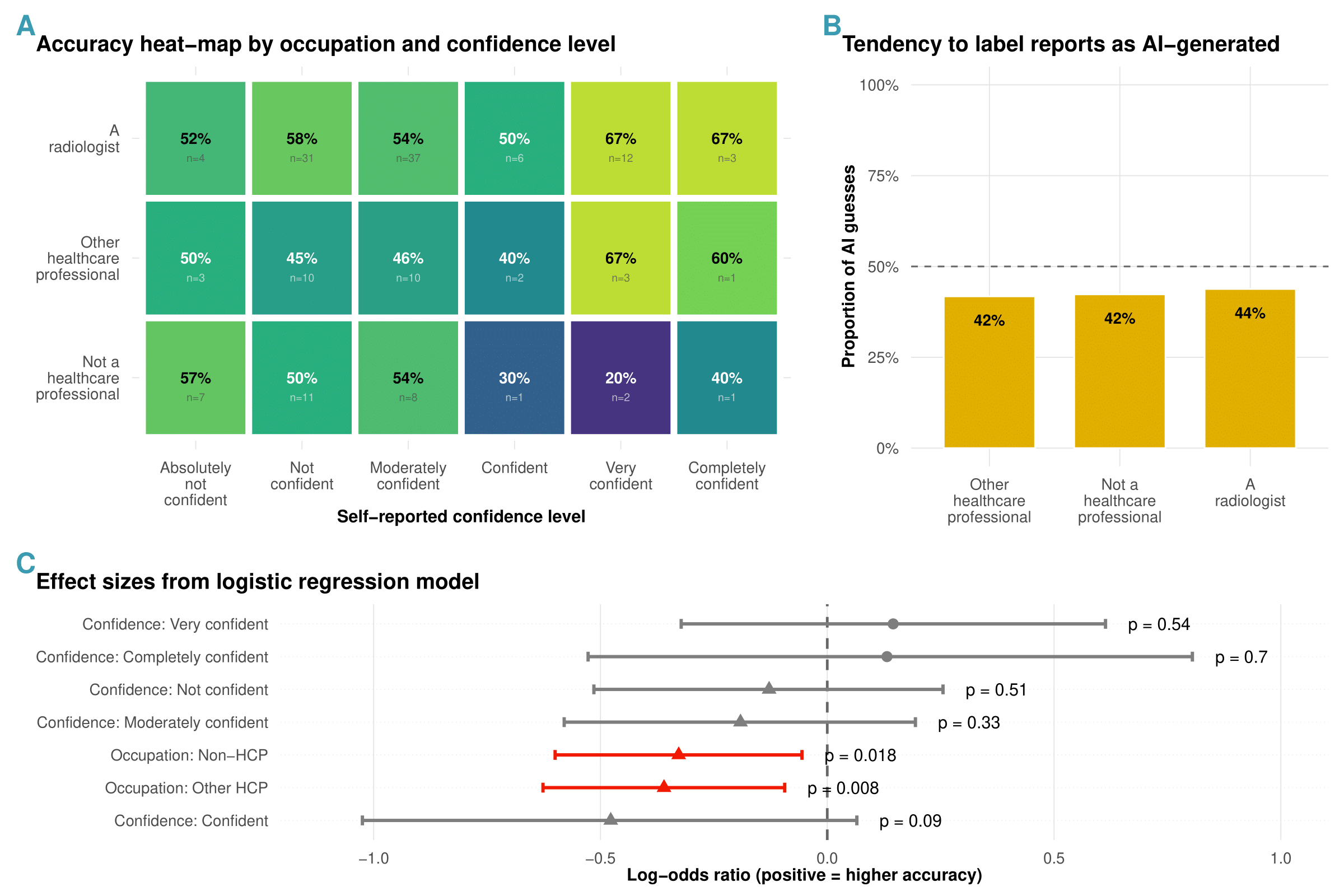} 
    \caption{Analysis of human vs. AI report differentiation performance A) Accuracy heat-map by occupation and confidence level, showing percentage accuracy for each combination with sample size indicated. Radiologists generally achieved higher accuracy across confidence levels compared to other groups. B) Tendency to label reports as AI-generated by occupation group, showing similar bias patterns across all groups with approximately 42-44\% of reports identified as AI-generated, indicating a slight bias towards labeling a report as human generated. C) Effect sizes from logistic regression model analyzing factors associated with discrimination accuracy. Non-healthcare professionals (Non-HCP) and other healthcare professionals (Other HCP) showed significantly lower accuracy compared to radiologists (p=0.018 and p=0.008 respectively), while confidence levels had varying but generally non-significant effects on performance.
}
    \label{fig:3}
\end{figure}

Analysis of response bias revealed a slight tendency across all professional groups to classify reports as human-authored rather than AI-generated. Radiologists classified \(43.8\%\) of reports as AI-generated, while non-healthcare professionals and other healthcare professionals classified \(42.3\%\) and \(41.8\%\), respectively (Figure~\ref{fig:2}). Logistic-regression analysis confirmed that professional background significantly influenced accuracy, with non-healthcare professionals (estimate: \(-0.328\), \(p = 0.018\)) and other healthcare professionals (estimate: \(-0.360\), \(p = 0.008\)) performing worse than radiologists (Figure~\ref{fig:3}).

\subsection{ICD-10 code distribution}

\begin{figure}[H]
    \centering
    \includegraphics[width=\textwidth,trim=0cm 0cm 0cm 0cm,clip,page=1]{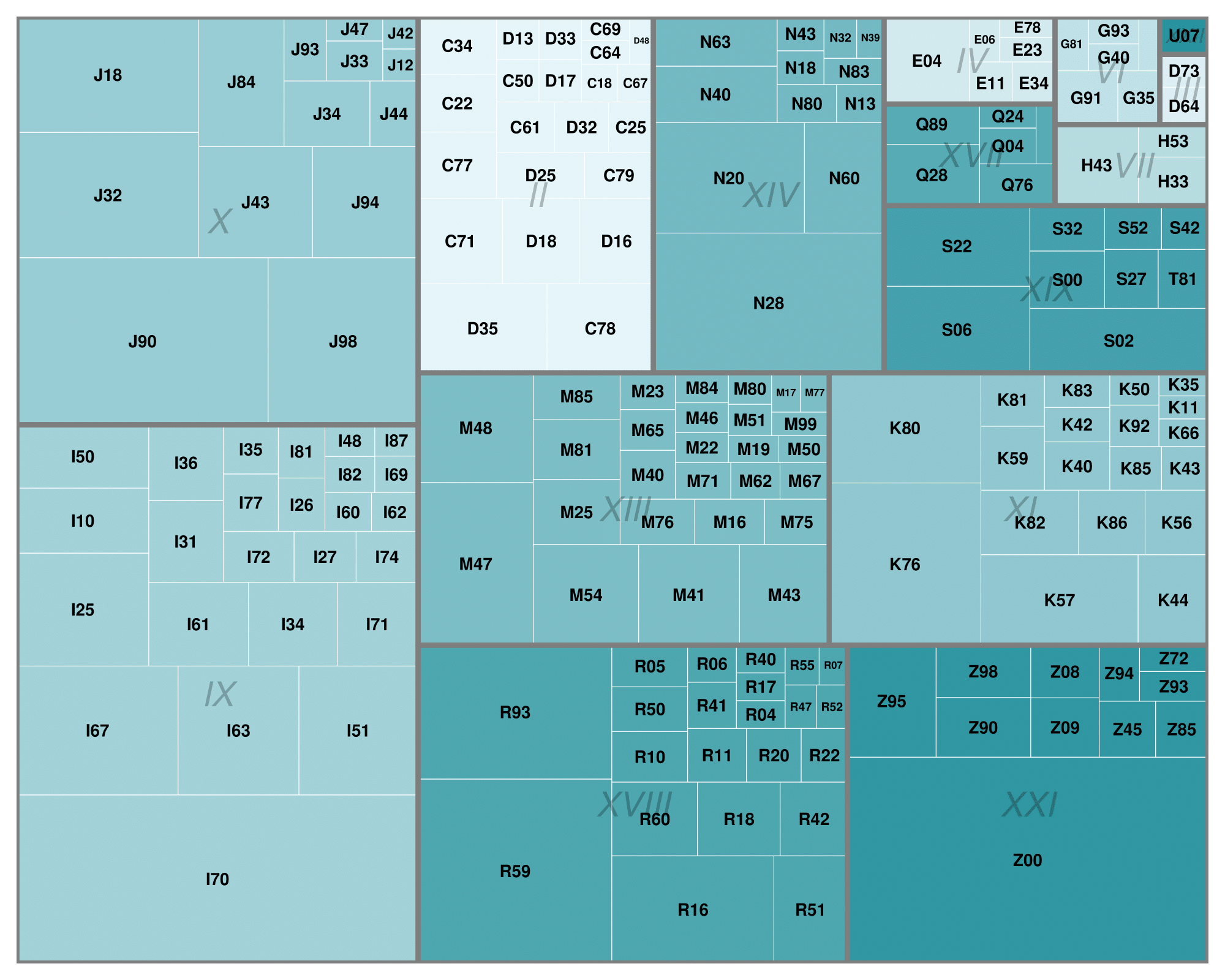} 
    \caption{Treemap of ICD-10 Codes in the PARROT dataset, grouped by the chapter number and 3 digit ICD-10 code. Refer to Table S1 for a detailed overview of each code and its subcodes. }
    \label{fig:4}
\end{figure}

The dataset encompasses pathologies across all major ICD-10 chapters, with notable predominance of codes from Chapter~IX (Diseases of the circulatory system), Chapter~X (Diseases of the respiratory system), Chapter~XI (Diseases of the digestive system), and Chapter~XIII (Diseases of the musculoskeletal system and connective tissue), which collectively account for \(53.9\%\) of all coded reports. Among specific three-digit codes, \texttt{I70} (Atherosclerosis) and \texttt{I63} (Infarction) represented the most frequent pathology categories, with \(133\) and \(109\) reports, respectively. Table~\ref{tab:3} provides an overview of the distribution of codes across chapters, and Figure~\ref{fig:4} presents a treemap illustrating the hierarchical distribution of codes grouped by chapter and three-digit classifications. 

\begin{table}[H]
\centering
\caption{Overview of ICD-10 codes}
\label{tab:3}
\footnotesize
\renewcommand{\arraystretch}{1.2}
\begin{tabularx}{\textwidth}{l>{\raggedright\arraybackslash}p{2cm}X}
\toprule
\textbf{Chapter} & \textbf{\% of Codes} & \textbf{3-Digit ICD-10 Codes} \\
\midrule
I   & 0.4  & A15, A18, A40, A41, B00, B16, B25, B37, B44, B58, B69, B96 \\
II  & 7.7  & C16, C18, C19, C20, C22, C23, C24, C25, C32, C34, C37, C41, C45, C50, C53, C54, C61, C64, C67, C69, C70, C71, C73, C74, C77, C78, C79, C81, C85, C90, C91, C92, D05, D10, D11, D13, D16, D17, D18, D24, D25, D28, D30, D31, D32, D33, D34, D35, D37, D41, D43, D45, D48 \\
III & 0.4  & D50, D57, D62, D64, D73, D76, D82, D86 \\
IV  & 1.5  & E03, E04, E06, E07, E11, E14, E22, E23, E27, E34, E46, E66, E78, E84, E85, E87 \\
IX  & 17.9 & I07, I10, I11, I12, I15, I21, I25, I26, I27, I28, I31, I33, I34, I35, I36, I37, I40, I42, I46, I48, I49, I50, I51, I60, I61, I62, I63, I64, I65, I66, I67, I69, I70, I71, I72, I73, I74, I77, I78, I80, I81, I82, I83, I87, I89, I95 \\
V   & 0.1  & F01, F03, F10, F33, F41, F79 \\
VI  & 1.5  & G00, G06, G20, G23, G25, G30, G31, G35, G36, G37, G40, G41, G43, G45, G50, G51, G54, G56, G57, G70, G81, G83, G90, G91, G93, G96, G97 \\
VII & 1.3  & H02, H04, H11, H20, H21, H27, H33, H35, H43, H44, H47, H53, H54 \\
VIII& 0.2  & H61, H65, H66, H70, H71, H83, H90, H92, H93 \\
X   & 13.8 & J01, J11, J12, J15, J18, J20, J21, J31, J32, J33, J34, J35, J36, J38, J39, J40, J42, J43, J44, J45, J47, J68, J69, J80, J81, J84, J86, J90, J92, J93, J94, J96, J98 \\
XI  & 8.9  & K02, K04, K08, K09, K11, K21, K22, K25, K26, K31, K35, K38, K40, K41, K42, K43, K44, K50, K51, K52, K55, K56, K57, K58, K59, K60, K63, K65, K66, K70, K74, K75, K76, K80, K81, K82, K83, K85, K86, K91, K92 \\
XII & 0.0  & L59, L72 \\
XIII& 9.8  & M06, M07, M11, M13, M15, M16, M17, M18, M19, M21, M22, M23, M24, M25, M34, M35, M40, M41, M43, M45, M46, M47, M48, M50, M51, M53, M54, M61, M62, M65, M67, M70, M71, M75, M76, M77, M79, M80, M81, M84, M85, M87, M89, M93, M94, M96, M99 \\
XIV & 7.4  & N10, N11, N13, N15, N17, N18, N19, N20, N21, N28, N30, N32, N39, N40, N41, N42, N43, N44, N50, N60, N61, N63, N64, N80, N82, N83, N84, N85, N89, N91, N92, N93, N94, N97 \\
XIX & 5.2  & S00, S02, S03, S05, S06, S12, S13, S20, S22, S26, S27, S32, S33, S36, S42, S43, S52, S53, S62, S70, S72, S73, S82, S83, S90, S92, S93, T17, T18, T44, T78, T80, T81, T82, T83, T88, T91, T92 \\
XV  & 0.1  & O00, O07, O43, O45 \\
XVI & 0.2  & P07, P21, P22, P24, P29, P35, P36, P78 \\
XVII& 2.2  & Q00, Q01, Q02, Q04, Q07, Q17, Q18, Q21, Q24, Q25, Q26, Q27, Q28, Q39, Q40, Q43, Q44, Q53, Q61, Q62, Q63, Q65, Q67, Q74, Q75, Q76, Q78, Q79, Q82, Q85, Q86, Q89, Q90 \\
XVIII& 11.7 & R00, R04, R05, R06, R07, R10, R11, R13, R16, R17, R18, R19, R20, R21, R22, R23, R25, R26, R27, R29, R30, R31, R33, R35, R40, R41, R42, R45, R47, R50, R51, R52, R55, R56, R58, R59, R60, R63, R64, R70, R74, R90, R93, R94 \\
XX  & 0.0  & V99 \\
XXI & 9.5  & Z00, Z01, Z03, Z04, Z08, Z09, Z31, Z34, Z42, Z45, Z47, Z72, Z80, Z85, Z90, Z92, Z93, Z94, Z95, Z96, Z98, Z99 \\
XXII& 0.1  & U07 \\
NA  & 0.0  & U09 \\
\bottomrule
\end{tabularx}
\end{table}

\section{Discussion}

In this work, we present \textbf{PARROT}, the largest openly available multilingual radiology-report dataset to date. Comprising \num{2658} fictional yet realistic reports contributed by 76 radiologists from 21 countries covering 13 languages, PARROT spans multiple imaging modalities (X-ray, ultrasound, CT, MRI) and nearly all major body regions. The collection's synthetic nature enables open sharing without privacy concerns while preserving authentic clinical language and style. By capturing diverse reporting practices across languages, modalities, and anatomical regions, PARROT serves as a resource for developing and evaluating NLP applications in a multilingual context.

Several open-access datasets for natural-language processing have been published in recent years, yet most remain limited in linguistic and geographic scope. The \textsc{MIMIC-IV} clinical database and its derivative \textsc{MIMIC-CXR} corpus represent among the largest collections of radiology reports, but they exclusively reflect English-language, U.S.-based practice \cite{johnson2023,johnson2019}. This monolinguistic bias affects research development: a systematic review found that, among 164 radiology-NLP studies, 142 (86.6 \%) focused exclusively on English reports, with minimal representation of other languages \cite{banerjee2024}.

Beyond language, openness and geographic diversity are critical considerations for dataset development. Many publicly available radiology-text datasets originate from a single country or healthcare system. The aforementioned collections, for instance, derive from U.S. hospital data \cite{johnson2023}. Privacy regulations governing clinical text—including HIPAA in the United States and GDPR in Europe—necessitate extensive de-identification protocols that often slow down or limit international data sharing. These constraints may lead to AI models being over-fitted to the phrasing, style, and clinical patterns of a limited context. Language and regional biases can therefore limit algorithm utility when confronted with different populations or reporting styles \cite{suenghataiphorn2025}. PARROT addresses these challenges by relying on fictional cases created by domain experts, offering a practical solution to data-accessibility issues.

The ongoing development of increasingly potent large language models, such as Generative Pre-trained Transformers (GPT), offers another route to synthetic radiology reports. By prompting frontier models like GPT-o1 \cite{o1card2024}, researchers could, in principle, generate artificial datasets tailored to specific use cases. However, this strategy may introduce subtle errors that are difficult to detect—especially for investigators without radiology expertise. Our differentiation experiment illustrates the risk: participants achieved only 53.9 \% accuracy overall in distinguishing PARROT reports from GPT-generated alternatives, with radiologists performing significantly better (56.9 \%, $p<0.05$) than other healthcare professionals (48.3 \%) or non-healthcare participants (49.7 \%). Because computer scientists conduct much of the NLP research, domain-specific errors in AI-generated reports might pass unnoticed \cite{wang2020}. PARROT offers expert-written text that embeds clinical-reasoning patterns current language models may not replicate fully, positioning it between purely synthetic corpora and real—but privacy-restricted—clinical records.

\subsection{Limitations}

Several limitations warrant consideration. First, the dataset is geographically imbalanced: Europe predominates, and only \(\approx\)19 \% of reports originate from the Global South. Broader participation is needed to capture additional languages, dialects, and regional reporting practices. Second, although fictional reports overcome privacy barriers, they lack links to real patient imaging and outcomes, limiting use in tasks that require image,text correlation or multimodal training. Recruitment through social media and professional networks may also introduce selection bias toward academically active contributors, potentially narrowing the range of clinical scenarios and styles represented. Finally, the current release includes only ICD-10 annotations; integrating domain-specific ontologies such as RadLex would enrich structured labels.

\subsection{Conclusion}

PARROT provides an open-access, multilingual collection of radiology reports that tackles linguistic and accessibility constraints hindering NLP tool development in non-English contexts. By offering fictional yet radiologist-authored reports across multiple languages, PARROT serves as a valuable benchmark for creating and testing AI applications capable of functioning across diverse healthcare systems and languages.

\nocite{*} 
\bibliography{main} 

\subsection{Supplementary figures}

\begin{figure}[H]
  \centering
  \includegraphics[width=\textwidth]{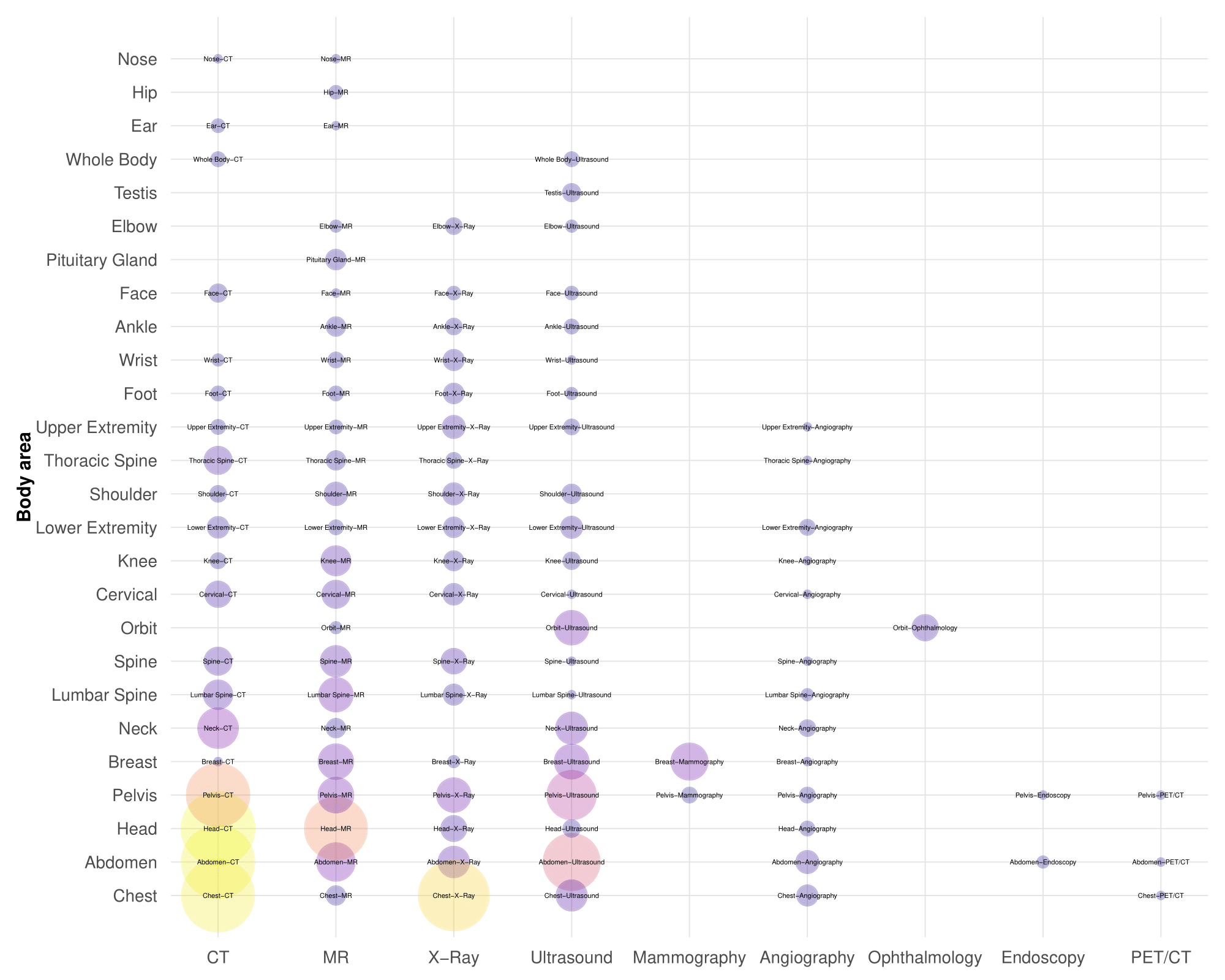}

  {\captionsetup{labelformat=empty,labelsep=none}
   \caption{\textbf{Supplementary Figure S1.}\;
            Bubble chart visualization displaying the frequency distribution of imaging examinations by anatomical region and modality. The bubble size and color intensity represent the number of reports for each combination, with larger, brighter bubbles indicating higher frequencies.}}
\end{figure}

\begin{figure}[H]
  \centering
  \includegraphics[width=\textwidth]{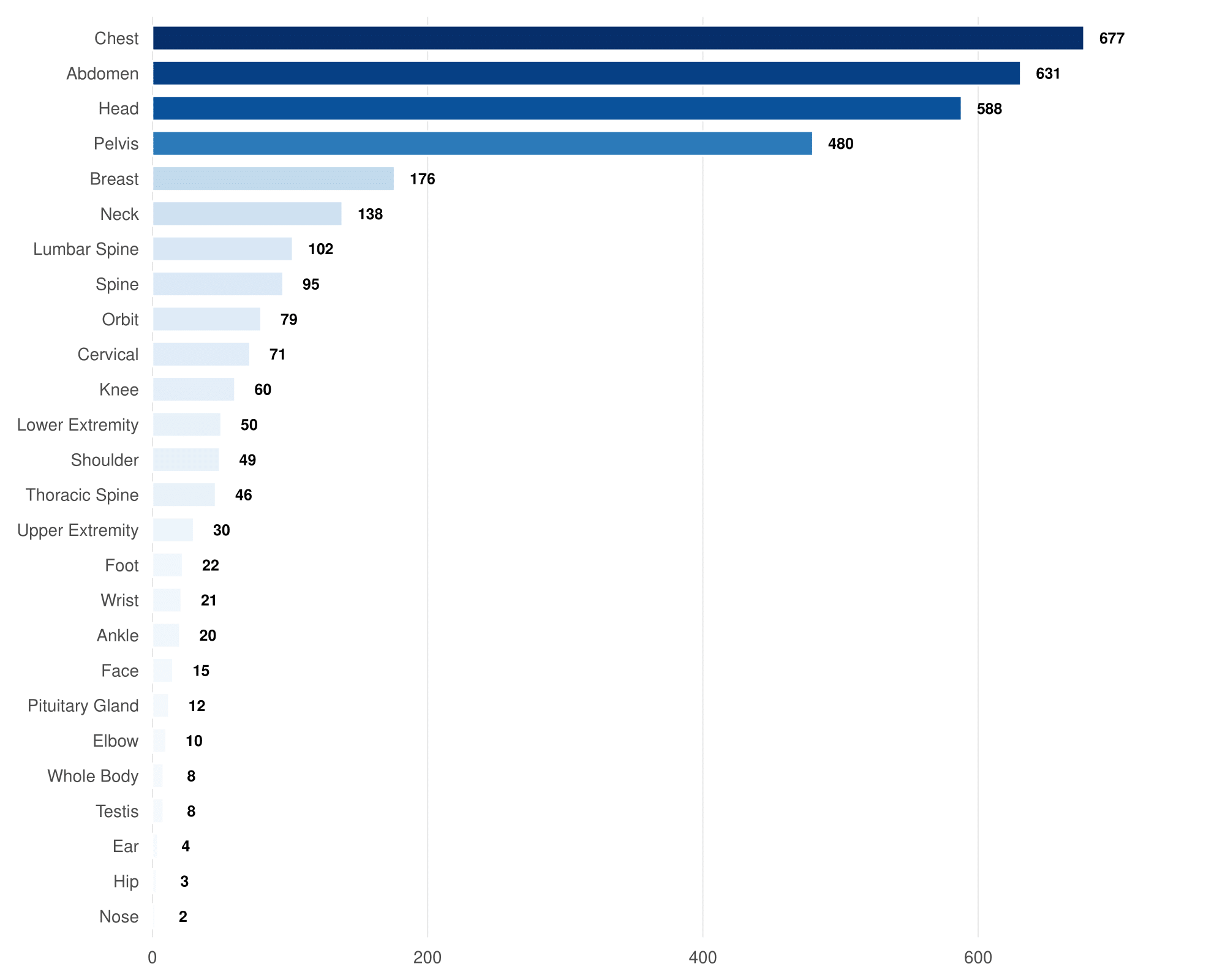}

  {\captionsetup{labelformat=empty,labelsep=none}
   \caption{\textbf{Supplementary Figure S2.}\;
            Horizontal bar chart displaying the frequency of anatomical regions represented in the dataset. Chest (677), Abdomen (631),Head (588), and Pelvis (480) are the predominant areas, accounting for approximately 70 \% of all reports.}}

\end{figure}

\end{document}